\documentclass[10pt,twocolumn,letterpaper]{article}

\usepackage{wacv}              

\usepackage{graphicx}
\usepackage{amsmath}
\usepackage{amssymb}
\usepackage{booktabs}
\usepackage{afterpage}
\usepackage{caption}
\usepackage{float}
\usepackage[accsupp]{axessibility}

\usepackage{multirow}
\usepackage{tabularx}
\usepackage{booktabs}
\usepackage[normalem]{ulem}
\useunder{\uline}{\ul}{}
\usepackage[pagebackref,breaklinks,colorlinks]{hyperref}
\usepackage{algorithm}
\usepackage{algorithmic}

\usepackage[capitalize]{cleveref}
\crefname{section}{Sec.}{Secs.}
\Crefname{section}{Section}{Sections}
\Crefname{table}{Table}{Tables}
\crefname{table}{Tab.}{Tabs.}

\def\wacvPaperID{1682} 
\def\confName{WACV}
\def\confYear{2025}

\begin{document}

\title{FlashVTG: Feature Layering and Adaptive Score Handling Network for Video Temporal Grounding}

\author{Zhuo Cao$^{1*}$, Bingqing Zhang$^1$\thanks{Equal Contribution}, Heming Du$^1$, Xin Yu$^1$, Xue Li$^1\thanks{Corresponding Authors}$, Sen Wang$^1$\\
\\ $^1$ {The University of Queensland, Australia} \\
{\tt\small \{william.cao, bingqing.zhang, heming.du, xin.yu\}@uq.edu.au}\\
{\tt\small xueli@eesc.uq.edu.au, sen.wang@uq.edu.au}
}

\maketitle

\begin{figure*}[!ht]
    \centering
    \includegraphics[width=\textwidth]{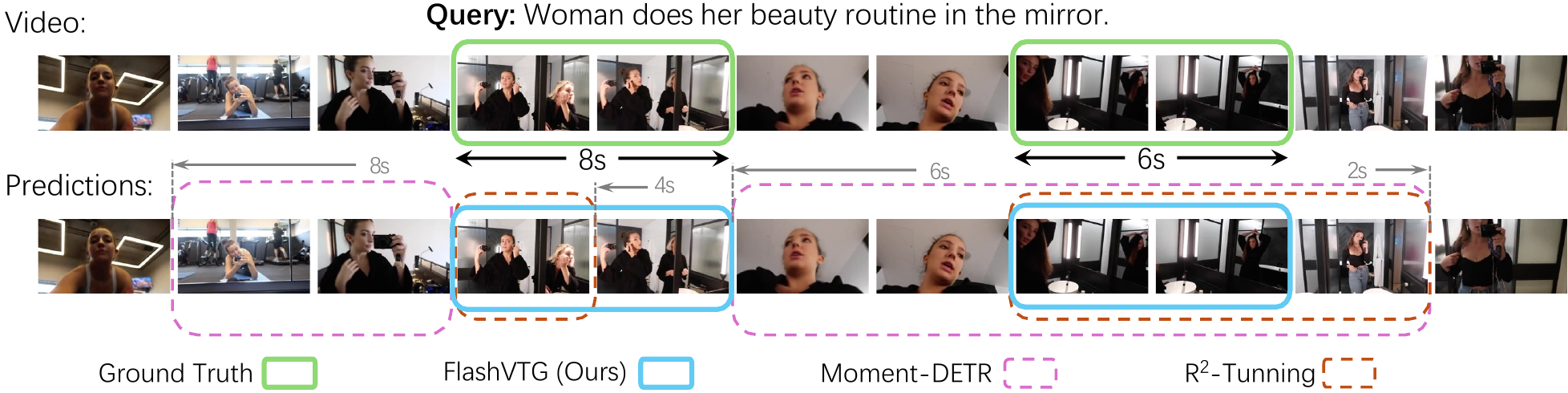}
    \caption{\textbf{Comparison of Model Performance on the Moment Retrieval Task} using video query pair from QVHighlights~\cite{lei2021detecting}. Ground Truth consists of two short moments, both of which are precisely retrieved by our model. In contrast, Moment-DETR \cite{lei2021detecting}, the established benchmark, and R2-Tuning \cite{liu2024tuning}, the previously leading method, failed to accurately retrieve the designated moments.}
    \label{fig:title fig}
\end{figure*}

\begin{abstract}
   Text-guided Video Temporal Grounding (VTG) aims to localize relevant segments in untrimmed videos based on textual descriptions, encompassing two subtasks: Moment Retrieval (MR) and Highlight Detection (HD). Although previous typical methods have achieved commendable results, it is still challenging to retrieve short video moments. This is primarily due to the reliance on sparse and limited decoder queries, which significantly constrain the accuracy of predictions. Furthermore, suboptimal outcomes often arise because previous methods rank predictions based on isolated predictions, neglecting the broader video context. To tackle these issues, we introduce FlashVTG, a framework featuring a Temporal Feature Layering (TFL) module and an Adaptive Score Refinement (ASR) module. The TFL module replaces the traditional decoder structure to capture nuanced video content variations across multiple temporal scales, while the ASR module improves prediction ranking by integrating context from adjacent moments and multi-temporal-scale features. Extensive experiments demonstrate that FlashVTG achieves state-of-the-art performance on four widely adopted datasets in both MR and HD. Specifically, on the QVHighlights dataset, it boosts mAP by 5.8\% for MR and 3.3\% for HD. For short-moment retrieval, FlashVTG increases mAP to 125\% of previous SOTA performance. All these improvements are made without adding training burdens, underscoring its effectiveness. Our code is available at https://github.com/Zhuo-Cao/FlashVTG.
\end{abstract}

\section{Introduction}
\label{sec:intro}
The increasing prevalence of video content across various platforms has amplified the need for advanced video analysis techniques, particularly in the context of Video Temporal Grounding (VTG). The task of VTG involves accurately identifying specific video segments that correspond to given natural language descriptions, a capability that is crucial for applications such as Moment Retrieval (MR), event detection, and Highlight Detection (HD). Addressing these challenges is critical as it directly impacts the performance and usability of systems that rely on video understanding. For instance, the ability to precisely localize and retrieve short yet significant moments within videos can enhance user experiences in applications ranging from video editing to automated video surveillance. Moreover, improving the ranking of the predictions ensures that the first retrieved moment is more accurate and contextually relevant, thereby reducing errors in downstream tasks.

Despite advancements in video temporal grounding, current methods remain limited, particularly in short moment retrieval in complex, densely packed video sequences. DETR~\cite{carion2020detr}-based models~\cite{lei2021detecting, moon2023qddetr, sun2024trdetr}, though effective, underperform in short moment retrieval due to their reliance on sparse and limited decoder queries, which tend to overlook short moments. However, simply increasing the number of queries significantly increases the computational complexity of the methods. Moreover, relying solely on isolated predicted moments for comparison and ranking can lead to suboptimal results, especially when fine-grained distinctions are required. 

In this paper, we introduce a Temporal Feature layering architecture (FlashVTG) to solve VTG tasks, including MR and HD. We first develop a Temporal Feature Layering module to extract and integrate video features across multiple temporal scales, enabling a more nuanced and comprehensive representation of video content. Subsequently, we introduce an Adaptive Score Refinement Module to rank predicted moments, enhancing the confidence scores of moments by integrating context from adjacent moments and multi-temporal-scale features. Together, these components enhance the ability to accurately predict moments of varying durations, particularly short moments (see Fig \ref{fig:title fig}), which are often problematic for previous methods.

Significant performance discrepancies are observed in DETR-based methods~\cite{moon2023cgdetr, jiang2024llmepet}, with mAP scores around 50\% for moments longer than 10 seconds and a drastic drop to approximately 10\% for shorter moments. These findings motivate us to develop a Temporal Feature Layering module for moment localization across various moment durations. We discard the conventional decoder structure and instead utilizes the Temporal Feature Layering module. By doing so, it addresses the inherent issue of sparse decoder queries in DETR-based methods, significantly improving the accuracy of moment retrieval without introducing additional training complexity.

For each predicted moment, a corresponding confidence score helps the model select the best prediction. However, we found that previous methods typically generate the confidence score only based on the current predicted moment, making it difficult to distinguish between closely adjacent moments when the video content is very similar. Therefore, we design the Adaptive Score Refinement Module. It enhances the model’s ability to generate accurate confidence scores by leveraging both intra-scale and inter-scale features. This adaptive mechanism evaluates predicted moments not only across different feature scales but also within adjacent predicted moments on the same scale. Additionally, to further enhance accuracy for short moments, we introduced a novel Clip-Aware Score Loss, which applies labels from the Highlight Detection task to the Moment Retrieval task, offering fine-grained supervision that was previously unexplored.

To validate the effectiveness of FlashVTG, we conducted extensive experiments on widely adopted VTG benchmarks. Integrating FlashVTG with InternVideo2~\cite{wang2024internvideo2}, SlowFast~\cite{feichtenhofer2019slowfast}, and CLIP~\cite{radford2021clip} yield remarkable results for moment retrieval and highlight detection.

These results not only underscore the robustness of FlashVTG but also demonstrate its superiority over state-of-the-art methods, with performance improvements of 2.7\%, 2.2\%, and 11.9\% respectively in MR, positioning FlashVTG as a leading approach in VTG tasks. Our contributions are as follows:
\vspace{-0.5em}
\begin{itemize}
    \item We propose FlashVTG, a novel architecture that significantly enhances Video Temporal Grounding by employing a strategic integration of Temporal Feature Layering and Adaptive Score Refinement.
    \vspace{-0.5em}
    \item We design a Temporal Feature Layering (TFL) Module that replaces the traditional decoder. TFL is designed to overcome sparse query limitations and improve retrieval without additional training.
    \vspace{-0.5em}
    \item We introduce an Adaptive Score Refinement (ASR) Module that selects predicted moments using both intra-scale and inter-scale features, enhancing first-moment accuracy and highlight detection.
\end{itemize}
\section{Related Work}
\label{sec:Related Work}

\begin{figure*}[t]
    \centering
    \includegraphics[width=0.8\linewidth]{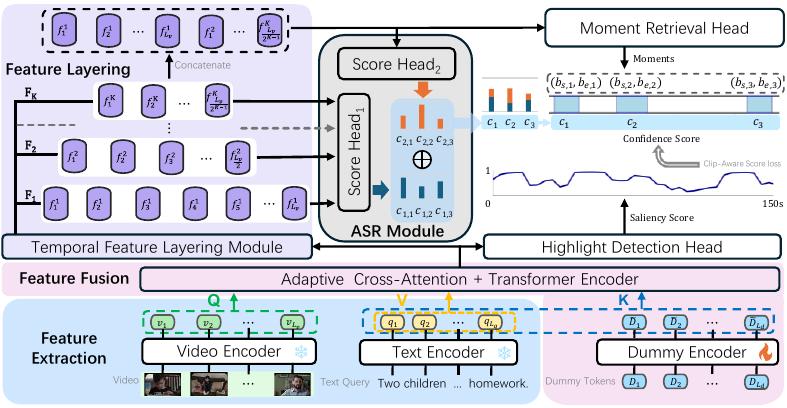}
    \caption{\textbf{Overview of the FlashVTG Framework.} As depicted in the blue section below, input videos and queries are first processed through frozen encoders to extract corresponding video and text features. These text features, concatenated with encoded Dummy tokens to form the Key, are merged with video text features in the Feature Fusion module to create Fused Features. These are then directed into the Temporal Feature Module and the HD Head, producing $K$ temporal scale features ($f^{i}_j$ refers to the token at the \(j^{th}\) position of the feature \(F_i\) at the \(i^{th}\) scale.) and saliency scores, respectively. These features and their concatenated forms are input into the Adaptive Score Refinement Module, generating intra- and inter-scale confidence scores $c$ (shown here with three examples). Lastly, the Moment Retrieval Head uses all fused features for boundary prediction, outputting timestamps $(b_s, b_e)$ for the start and end points.}
    \label{fig:overview}
\end{figure*}

\textbf{Video Temporal Grounding (VTG).} In the field of text-guided Video Temporal Grounding~\cite{lei2021detecting, lin2023univtg, jiang2024llmepet}, the objective is to identify specific temporal segments within a video based on given natural language descriptions. Essentially, this task requires models to discern and quantify the associations between the content of the video and natural language descriptions across different time stamps. Subsequently, we will provide a detailed exposition of two specific VTG tasks: moment retrieval and highlight detection.

\textbf{Moment Retrieval (MR).} The goal of this task is to identify the start and end points of a video segment based on a natural language query. Given the limitations of current datasets, it is typically assumed that a single ground truth (GT) segment exists. If multiple GT segments are available, the one with the highest Intersection Over Union (IOU) with the predicted moment is selected as the GT. Existing MR methods primarily fall into two categories: proposal-based~\cite{zhang2020learning, xu2019multilevel, xiao2021boundary, shao2018find, liu2021context, zhang2019cross, liu2020jointly} and proposal-free methods~\cite{liu2022reducing, lu2019debug, chen2020rethinking, zeng2020dense, mun2020local, lei2021detecting, liu2022memory}.

Proposal-based methods operate in two stages: generating candidate moments and treating retrieval as a matching problem. These methods can be further classified into sliding window~\cite{yuan2019semantic, liu2018attentive, ge2019mac, anne2017localizing, gao2017tall}, anchor-based~\cite{zhang2019cross, yuan2019semantic, wang2020temporally, zhang2019man}, and proposal-generated~\cite{zhang2020learning, xu2019multilevel, xiao2021boundary, shao2018find, liu2021context} approaches. In contrast, proposal-free methods~\cite{liu2022memory} use regression to directly predict the start and end points of relevant moments.

\textbf{Highlight Detection (HD).} Similar to MR, HD aims to assess the importance of video segments, but here it involves determining the relevance of each segment to a given query. Each clip is assigned a score reflecting its correlation with the query. From a methodological perspective, early research on HD predominantly utilized ranking-based approaches~\cite{gygli2016video2gif, liu2011query, rochan2020adaptive, sun2014ranking, liu2015multi}. In recent years, methods~\cite{lei2021detecting, moon2023qddetr, moon2023cgdetr, jang2023eatr} based on the DETR have exhibited notable performance improvements. 

Although both MR and HD are closely related, they were not studied concurrently until the publication of the QVHighlights dataset~\cite{lei2021detecting}, which prompted numerous works integrating these tasks. The first study to merge them, Moment-DETR~\cite{lei2021detecting}, was introduced alongside the QVHighlights dataset. This model serves as a baseline capable of addressing both MR and HD challenges. Subsequent research has built on this framework with various enhancements: UMT~\cite{liu2022umt} incorporates audio information into query generation; QD-DETR~\cite{moon2023qddetr} enhances performance through negative video-text pair relationships; UVCOM~\cite{xiao2024uvcom} improves outcomes by considering the varying importance of different tasks at different granularities; CG-DETR~\cite{moon2023cgdetr} enhances alignment between queries and moments by considering the clipwise cross-modal interactions; and LLMEPET~\cite{jiang2024llmepet} integrates LLM encoders into existing Video Moment Retrieval (VMR) architectures. Beyond DETR-based methods, the recent $R^{2}$-Tunning~\cite{liu2024tuning} framework, which leverages transfer learning from images to videos, has shown potential capabilities for the VTG task. However, these methods struggle to handle multi-scale information, particularly for short moment retrieval, leading to performance bottlenecks. in FlashVTG, we process multi-scale information separately, achieving improvements in both MR and HD tasks.

\section{Methodology}
\label{sec:method}
\subsection{Problem Formulation}
In Video Temporal Grounding (VTG) tasks, we represent the input video and the extracted features as $\mathcal{V}$ and $\mathbb{V} = \{v_i\}_{i=1}^{L_v}$, respectively, where each $v_i \in \mathbb{R}^{D_v}$. 
Here, $L_v$ denotes the number of video clips and $D_v$ represents the feature dimension of each clip. For the natural language query $\mathcal{Q}$, we represent it as a set $\mathbb{Q} = \{q_i\}_{i=1}^{L_q}$, where each $q_i \in \mathbb{R}^{D_q}$ is the $i$-th word token in the query. Here, $L_q$ denotes the word number of query and $D_q$ indicates the feature dimension of each word token.

For Moment Retrieval (MR), the objective of the model is to predict the start point $b_s$ and the end point $b_e$ of the moment based on the input $V$ and the query. In the common practices, the output is a set of tuples $(b_s, b_e)$ representing the temporal boundaries of these moments. Additionally, the model generates confidence scores $c$ to rank these predictions, although these scores are not required by the task itself. Thus, the model produces: $\{(b_{s,i}, b_{e,i}, c_i) \mid i = 1, 2, \ldots, n\}$, with confidence scores $\{c_1, c_2, \dots, c_n\}$ used for ranking.

In the Highlight Detection task, the model outputs a sequence of clip-wise saliency scores $S = \{s_i\}_{i=1}^{L_v}$, where each score $s_i \in [0,1]$ quantifies the relevance of the respective clip to the query.

\subsection{Overall Framework}
As illustrated in Figure \ref{fig:overview}, the input video and query are initially encoded into video and query features using frozen Feature Encoders. In addition to extracting the original video and text features, FlashVTG incorporates a dummy token to supplement semantic information beyond the query, enhancing alignment between the video and text.

Following the encoding process, the video and query features are integrated within the Feature Fusion Module, enabling precise alignment of the two modalities. The fused features are subsequently processed in two separate streams. For Highlight Detection, the HD head transforms the fused features into saliency scores, while the Temporal Feature Layering Module expands the features across multiple granularities.

The Adaptive Score Refinement Module then evaluates these multi-scale features to generate a confidence score, $s$. Concurrently, the Moment Retrieval prediction head outputs predictions across different scales, producing a series of tuples $(b_s, b_e)$. Finally, the predicted moments and confidence scores are combined into ${(b_{s,i}, b_{e,i}, c_i) \mid i = 1, 2, \ldots, n}$, which specify the predicted start and end points of moments, along with their associated confidence scores.

\subsection{Feature Extraction and Fusion Module}
\label{subsec: Feature Fusion Module}
\textbf{Feature Extraction}. Consistent with previous methodologies, we employ the CLIP~\cite{radford2021clip} image encoder and SlowFast~\cite{feichtenhofer2019slowfast} to extract clip-level video features $\mathbb{V}$, and use the CLIP text encoder along with GloVe~\cite{pennington2014glove} to derive word-level text features $\mathbb{Q}$. Specifically, raw videos are segmented into clips at predetermined FPS, such as 0.5 or 1, using frozen pretrained models as feature extractors. This process transforms each clip into a distinct video feature $v_i$. Similarly, each word in the query is encoded into corresponding text features $q_i$.

\textbf{Feature Fusion.} To align video and query features for subsequent processing, we project them into the same dimensional space $d$ using two MLPs before fusion. We adopt the Adaptive Cross Attention (ACA) module from CG-DETR~\cite{moon2023cgdetr}, which extends the traditional Cross Attention by incorporating the ability to supplement query information. A known limitation of standard Cross Attention, particularly with the softmax operation, is that it may not perfectly align all video clips with the query, as the query cannot fully encapsulate the semantic scope of an untrimmed video. In ACA module, we use learnable dummy tokens, $D = [D_1, ..., D_{L_d}]$, where $L_d$ is a hyperparameter. These tokens capture semantic information beyond the original query, complementing its content. The encoded dummy tokens $\tilde{D}$ are concatenated with the query features $\mathbb{Q}$ to form the Key, with the video features $\mathbb{V}$ as the Query and the query features $\mathbb{Q}$ as the Value, as shown in \Cref{eq QKV}.
\begin{equation}
    \begin{aligned}
        \label{eq QKV}
        \textit{Query} &= [p_Q(v_1), \ldots, p_Q(v_{L_v})] \\
        \textit{Key} &= [p_K(q_1), \dots, p_K(q_{L_q}), p_K(\tilde{D_1}), \dots, p_K(\tilde{D_{L_d}})] \\
        \textit{Value} &= [p_V(q_1), \dots, p_V(q_{L_q})]
    \end{aligned}
\end{equation}

Here, \(p_Q(\cdot)\), \(p_K(\cdot)\), and \(p_V(\cdot)\) denote the projection functions used to transform inputs into the Query, Key, and Value formats, respectively. These Query, Key, Value are input into the ACA module to obtain the fused features $ F \in \mathbb{R}^{L_v \times d} $, as shown in \Cref{eq:fused feature,eq:W}.
\begin{align}
    \textit{F} = \text{ACA}(v_i) = \sum_{j=1}^{L_q} W_{i,j} \odot \textit{V}_j; \label{eq:fused feature} \\
    W_{i,j} = \frac{
        \text{exp}\left( \frac{\textit{Q}_i \odot \textit{K}_j}{\sqrt{d}} \right)}
        {\sum_{k=1}^{L_q+L_d} \text{exp}\left( \frac{\textit{Q}_i \odot \textit{K}_k}{\sqrt{d}} \right)}, \label{eq:W}
\end{align}
where $F$ denotes the fused feature and $\odot$ stands for the dot product. While $Q$, $K$, and $V$ represent the Query, Key, and Value, respectively. Through this adaptive cross-attention mechanism, the model more effectively aligns the query with the relevant video segments.

Following the ACA module, the fused features $F$ are passed through a Transformer Encoder to further refine the multi-modal representations. This additional processing step enables the model to enhance the interaction between the video and query features, allowing for a more comprehensive understanding of the temporal and semantic relationships across the entire sequence. 


\textbf{Highlight Detection Head}. The refined features output from the Transformer Encoder are further processed to generate clip-level saliency scores~$s$ for highlight detection. We aggregate the fused features to form a global contextual representation, which is then combined with the original fused features through a linear projection. The operation can be expressed as: 

\begin{equation}
s = \frac{\sum_{i} \left( \textbf{W}_1 \textit{f}_i \circ \textbf{W}_2 \textit{g} \right)}{\sqrt{d}}, 
\end{equation}
where $\textbf{W}_1$ and $\textbf{W}_2$ represent two linear projection functions, $\textit{f}_i$ denotes the $i$-th clip-wise fused features, and $\textit{g}$ represents the global contextual features. The symbol $\circ$ represents the Hadamard operation. This operation identifies the key clips in the video, yielding a sequence of scores that quantify the relevance of each clip to the overall content.

\subsection{ Temporal Feature Layering Module}
\label{subsec: Temporal Feature layering Module}
\textbf{Temporal Feature Layering.} This module transforms the fused features into a feature pyramid to enhance the model's capability to process features at various granularities. The feature pyramid is a commonly used structure in computer vision for extracting and utilizing vision features across multiple scales~\cite{wang2021pyramid,ghiasi2019fpn}. This architecture emulates the hierarchical processing of the human visual system, while enabling the network to effectively perceive and process video information of varying lengths. 

Building on this insight, we incorporate layered feature implementations into DETR-based models to overcome their limitations in short moment retrieval. This module separates moments of varying lengths, enabling the subsequent ASR module (Sec.~\ref{subsec: Adaptive Score Refinement Module}) to provide more detailed supervision for short moments prediction. Specifically, we apply 1D convolution operations with various strides to the fused features $F$, constructing a feature pyramid composed of features at different granularities. Predictions of moment locations are then made at these various granularities using a uniform prediction head.

Let \( F \in \mathbb{R}^{L_v \times d} \) represent the input fused feature, where \( L_v \) is the length of the features and \( d \) is the dimensionality of the features. The process of temporal feature layering can be represented as:
\begin{align*}
    F_{k} = 
    \begin{cases}
        F, & \text{if } k=1, \\
        \text{Conv1D}^{k-1}(F, \text{stride}=2), & \text{if } k=2, 3, \dots, K.
    \end{cases}
    \label{eq:feature pyramid}
\end{align*}

By applying the temporal feature layering, we obtain a sets of features at different granularities, which can be represented as:
\begin{align*}
    F_k \in \mathbb{R}^{\frac{L_v}{2^{k-1}} \times d}, \quad k = 1, 2, \dots, K.
\end{align*}

Here, $\{ F_k | k=1, 2, \dots, K\}$ represent the feature pyramid obtained from the original fused feature \( F \) through multiple convolution operations. This multi-scale processing approach enables the model to capture and process information at different temporal resolutions, thereby adapting to scene changes at various scales. As referenced in Sec.~\ref{sunsec: ablation study}, this module significantly improves the model's ability to retrieve moments of varying lengths. This operation paves the way for enhanced supervision of short moment retrieval in subsequent steps.

\textbf{Moment Prediction Head.} This module primarily adjusts and reduces the feature dimension to 2, and through a series of transformations, it yields the predicted start and end points of moments. The specific process can be expressed as:
\begin{equation}
B_k = \left( \sigma \left( \text{Conv1D}\left( \sigma \left( \text{Conv1D}(F_k) \right) \right)^\top \right) \right)^\top \times C_k.
\end{equation}
Here, \( B_k \in \mathbb{R}^{\frac{L_v}{2^{k-1}} \times 2} \) represents a set of boundaries at scale \( k \), \( \sigma (\cdot) \) denotes the ReLU activation function. The term \( C_k \) refers to a learnable parameter corresponding to each scale, which is used to adjust the influence of different scales on boundary prediction.

\begin{table*}[ht]
\centering
\small
\setlength{\tabcolsep}{3mm}{
\begin{tabular}{@{}lcccccccccc@{}}
\toprule
\multirow{3}{*}[-0.5em]{\textbf{Method}} & \multicolumn{5}{c}{test} & \multicolumn{5}{c}{val}  \\ \cmidrule(lr){2-6} \cmidrule(l){7-11} 
 & \multicolumn{2}{c}{R1}   & \multicolumn{3}{c}{mAP}  & \multicolumn{2}{c}{R1} & \multicolumn{3}{c}{mAP} \\ 
\cmidrule(lr){2-3} \cmidrule(lr){4-6} \cmidrule(lr){7-8} \cmidrule(l){9-11}  
 & @0.5 & @0.7   & @0.5   & @0.75  & Avg.   & @0.5  & @0.7  & @0.5  & @0.75 & Avg. \\ \midrule
M-DETR \cite{lei2021detecting} {\scriptsize \textit{NeurIPS'21}} & 52.89  & 33.02  & 54.82  & 29.17  & 30.73  & 53.94 & 34.84 & -  & -  & 32.20   \\
UMT \cite{liu2022umt} {\scriptsize \textit{CVPR'22}} & 56.23  & 41.18 & 53.83  & 37.01  & 36.12  & 60.26 & 44.26 & 56.70 & 39.90 & 38.59   \\
QD-DETR \cite{moon2023qddetr} {\scriptsize \textit{CVPR'23}}  & 62.40  & 44.98 & 62.52  & 39.88  & 39.86  & 62.68 & 46.66 & 62.23 & 41.82 & 41.22   \\
UniVTG \cite{lin2023univtg} {\scriptsize \textit{ICCV'23}} & 58.86  & 40.86 & 57.60  & 35.59  & 35.47  & 59.74 & -  & -  & -  & 36.13   \\
EaTR \cite{jang2023eatr} {\scriptsize \textit{ICCV'23}} & - & -  & - & - & - & 61.36 & 45.79 & 61.86 & 41.91 & 41.74   \\
MomentDiff \cite{li2024momentdiff}  {\scriptsize \textit{NeurIPS'23}} & 57.42  & 39.66 & 54.02  & 35.73  & 35.95  & -  & -  & -  & -  & - \\
TR-DETR \cite{sun2024trdetr} {\scriptsize \textit{AAAI'23}}  & 64.66   & 48.96 & 63.98  & 43.73  & 42.62  & 67.10 & 51.48 & 66.27 & 46.42 & 45.09   \\
TaskWeave \cite{yang2024taskweave} {\scriptsize \textit{CVPR'24}}  & - & -  & - & - & - & 64.26 & 50.06 & 65.39 & 46.47 & 45.38   \\
CG-DETR \cite{moon2023cgdetr} {\scriptsize \textit{Arxiv'24}} & 65.43 & 48.38 & 64.51  & 42.77  & 42.86  & 67.35 & 52.06 & 65.57 & 45.73 & 44.93   \\
UVCOM \cite{xiao2024uvcom} {\scriptsize \textit{CVPR'24}}  & 63.55   & 47.47 & 63.37  & 42.67  & 43.18  & 65.10 & 51.81 & -  & -  & 45.79   \\
SFABD \cite{huang2024semantic}  {\scriptsize \textit{WACV'24}} & - & - & 62.38 & 44.39 & 43.79  & - & - & - & - & -\\
LLMEPET \cite{jiang2024llmepet} {\scriptsize \textit{MM'24}} & 66.73   & 49.94 & 65.76  & 43.91  & 44.05  & 66.58 & 51.10 & -  & -  & 46.24   \\
$R^2$-Tunning \cite{liu2024tuning} {\scriptsize \textit{ECCV'24}}  & \underline{68.03} & 49.35 & \underline{69.04}  & 47.56 & 46.17  & 68.71 & 52.06 & -  & -  & 47.59   \\
\midrule
\textbf{FlashVTG} (Ours) & 66.08   & \underline{50.00}  & 67.99 & \underline{48.70}  & \underline{47.59}  & \underline{69.03}   & \underline{54.06} & \underline{68.44} & \underline{52.12}  & \underline{49.85} \\ 
\textbf{FlashVTG}\textsuperscript{\dag} \text{(Ours)} & \textbf{70.69} & \textbf{53.96}  & \textbf{72.33}  & \textbf{53.85}  & \textbf{52.00}  & \textbf{73.10}   & \textbf{57.29}   & \textbf{72.75} & \textbf{54.33}  & \textbf{52.84}  \\
\bottomrule
\end{tabular}
}
\caption{Performance comparison on the QVHighlights~\cite{lei2021detecting} Test and Validation Splits. In each column, the highest score is highlighted in \textbf{bold}, and the second highest score is \underline{underlined}. The notation \dag~indicates that the backbone used is InternVideo2~\cite{wang2024internvideo2}.}
\label{tab:qv_mr}
\end{table*}

\subsection{Adaptive Score Refinement Module}
\label{subsec: Adaptive Score Refinement Module}
Adaptive Score Refinemen Module used to assign a confidence score $c\in[0,1]$ to each predicted moment in MR. This score indicates the extent to which the given query is relevant to the predicted boundary. 
Compared with the previous method of generating scores on a single scale feature, we leverage both intra-scale and inter-scale scores to improve the final predictions. 

For each level of the feature pyramid $F_k$, intra-scale scores are first generated through a score head, producing outputs corresponding to the varying dimensions of the pyramid levels. These outputs are then concatenated into a unified tensor along the length dimension. These steps are shown in \Cref{eq:scorehead1} and \Cref{eq:score1}, respectively.
\begin{align}
    &c_k = \text{ScoreHead}_1(F_k) \in \mathbb{R}^{\frac{L_v}{2^{k-1}} \times 1}, k=1, 2, \dots, K.  \label{eq:scorehead1}\\
    &c_{\text{intra}} = \text{Concat}(c_1, c_2, \dots, c_K). \label{eq:score1}
\end{align}

Our score head utilizes a 2D convolutional network with a kernel size of \(1 \times 5\), which is effectively equivalent to a 1D convolution. Simultaneously, as shown in \Cref{eq: score2}, inter-scale scores are computed by concatenating the features from all pyramid levels and passing them through another score head, resulting in a tensor that matches the dimensions of the intra-scale output. 
\begin{align}
    &c_{\text{inter}} = \text{ScoreHead}_2(\text{Concat}(F_1, F_2, \dots, F_K)). \label{eq: score2}
\end{align}

The final prediction is obtained by a weighted combination of the intra-scale and inter-scale scores:
\begin{equation}
    c_{\text{final}} = x \cdot c_{\text{intra}} + (1 - x) \cdot c_{\text{inter}}.
\end{equation}

Here, the learnable weighting factor $x$ enables adaptive adjustment between $c_{\text{intra}}$ and $c_{\text{inter}}$, thereby resulting in a more comprehensive score prediction.

\subsection{Training Objectives}
\label{subsec: Training Objectives}
In FlashVTG, we employ a series of loss functions to ensure the model converges towards the desired objectives. For MR, we use Focal Loss~\cite{lin2017focal}, L1 Loss, and Clip-Aware Score Loss to respectively optimize the classification labels, boundaries, and clip-level confidence scores of the predicted moments. For HD, we utilize SampledNCE Loss~\cite{liu2024tuning} and Saliency Loss to optimize the saliency scores for each clip. The overall loss can be expressed as:
\begin{align*}
\mathcal{L}_{\text{overall}} &= \lambda_{\text{Reg}} \mathcal{L}_{\text{L1}} + \lambda_{\text{Cls}} \mathcal{L}_{\text{Focal}} + \lambda_{\text{CAS}} \mathcal{L}_{\text{CAS}}  \\
&\quad + \lambda_{\text{SNEC}} \mathcal{L}_{\text{SNCE}} + \lambda_{\text{Sal}} \mathcal{L}_{\text{Sal}},
\end{align*}
where each $\lambda_*$ represents the corresponding weight for each loss component. Due to the space limitation, we will focus on the Clip-Aware Score Loss, further details on the other loss functions can be found in the supplementary materials.

\textbf{Clip-Aware Score Loss.} This Loss is designed to align the predicted confidence scores with the target saliency labels. Given the predicted clip-wise moment confidence score $c_{\text{final}}$ and the target saliency scores $s_{\text{gt}}$, we first normalize both sets of scores using min-max normalization to get $\hat{c}_{\text{final}}$ and $\hat{s}_{\text{gt}}$. The loss is then computed as the mean squared error between $\hat{c}_{\text{final}}$ and $\hat{s}_{\text{gt}}$:
\begin{equation}
\mathcal{L}_{\text{CAS}} = \text{MSE}(\hat{c}_{\text{final}}, \hat{s}_{\text{gt}}).
\end{equation}

This loss encourages the model to produce confidence scores that not only match the target labels but also adhere to the relative distribution of saliency across clip-level, thereby specifically enhancing the model's performance when predicting short moments.

\begin{table}
\centering
\small
\begin{tabular}{lcccc}
\toprule
{\textbf{Method}}                   & R@0.3           & R@0.5           & R@0.7           & mIoU            \\ 
\cmidrule(lr{0.25em}){1-5}
2D-TAN \cite{zhang2020learning}        & 40.01           & 27.99           & 12.92           & 27.22           \\
VSLNet \cite{zhang2020vslnet}       & 35.54           & 23.54           & 13.15           & 24.99           \\
Moment-DETR \cite{lei2021detecting} & 37.97           & 24.67           & 11.97           & 25.49           \\
UniVTG \cite{lin2023univtg}         & 51.44           & 34.97           & 17.35           & 33.60           \\
CG-DETR \cite{moon2023cgdetr}       & 52.23           & \underline{39.61} & 22.23           & 36.48           \\
$R^2$-Tuning \cite{liu2024tuning}   & 49.71           & 38.72           & \textbf{25.12}  & 35.92           \\
LLMEPET \cite{jiang2024llmepet}     & \underline{52.73} & -               & 22.78           & \underline{36.55} \\ 
\cmidrule(lr{0.25em}){1-5}
\textbf{FlashVTG} \text{(Ours)}            & \textbf{53.71}  & \textbf{41.76}  & \underline{24.74} & \textbf{37.61}  \\ 
\bottomrule
\end{tabular}%
\caption{Performance Evaluation on TACoS~\cite{regneri2013tacos}. All these methods utilize SlowFast~\cite{feichtenhofer2019slowfast} and CLIP~\cite{radford2021clip} as backbones for TACoS. The highest score in each column is \textbf{bolded}, and the second highest is \underline{underlined}.}
\label{tab:tacos}
\vspace{-1.5em}
\end{table}

\begin{table}[!ht]
    \centering
    \small
    \begin{tabular}{lccc}
        \toprule
        \textbf{Method} & Backbone & R1@0.5 & R1@0.7 \\
        \midrule
        SAP~\cite{chen2019SAP} & VGG & 27.42 & 13.36 \\
        TripNet~\cite{hahn2019tripnet} & VGG & 36.61 & 14.50 \\
        MAN~\cite{zhang2019man} & VGG & 41.24 & 20.54 \\
        2D-TAN~\cite{zhang2020learning} & VGG & 40.94 & 22.85 \\
        FVMR~\cite{li2021fvmr} & VGG & 42.36 & 24.14 \\
        UMT\textsuperscript{\dag}~\cite{liu2022umt} & VGG & 48.31 & 29.25 \\
        QD-DETR~\cite{moon2023qddetr} & VGG & 52.77 & 31.13 \\
        TR-DETR~\cite{sun2024trdetr} & VGG & 53.47 & 30.81 \\
        CG-DETR~\cite{moon2023cgdetr} & VGG & \textbf{55.22} & \underline{34.19}  \\
        \cmidrule(lr{0.25em}){1-4}
        \textbf{FlashVTG} (ours) & VGG & \underline{54.25} & \textbf{37.42} \\
        \midrule
        2D-TAN \cite{zhang2020learning}  & SF+C & 46.02 & 27.50 \\
        VSLNet \cite{zhang2020vslnet} & SF+C      & 42.69           & 24.14           \\
        Moment-DETR \cite{lei2021detecting} & SF+C & 52.07           & 30.59           \\
        QD-DETR \cite{moon2023qddetr} & SF+C      & 57.31           & 32.55           \\
        UniVTG \cite{lin2023univtg}   & SF+C      & 58.01           & 35.65           \\
        TR-DETR~\cite{sun2024trdetr} & SF+C & 57.61 & 33.52 \\
        CG-DETR \cite{moon2023cgdetr} & SF+C      & \underline{58.44}           & 36.34           \\
        LLMEPET \cite{jiang2024llmepet}  & SF+C   & -               & \underline{36.49}           \\ 
        \cmidrule(lr{0.25em}){1-4}
        \textbf{FlashVTG} \text{(Ours)} & SF+C & \textbf{60.11} & \textbf{38.01} \\
        \textbf{FlashVTG} \text{(Ours)} & IV2 & \textbf{70.32} & \textbf{49.87} \\
        \bottomrule
    \end{tabular}
    \caption{Experimental results on the Charades-STA test set. ``SF+C" refers to SlowFast R-50~\cite{feichtenhofer2019slowfast} combined with CLIP-B/32~\cite{radford2021clip}, and ``IV2" denotes InternVideo2-6B~\cite{wang2024internvideo2}.c Methods marked with "†" use audio features.}
    \label{tab:Charades-STA}
    \vspace{-2em}
\end{table}
\vspace{-1.5mm}
\section{Experiments}
\label{sec:experiment}
\subsection{Datasets}
We evaluated our model on five VTG task datasets, including QVHighlights, TACoS, Charades-STA, TVSum, and YouTube-HL.

QVHighlights~\cite{lei2021detecting} is the most widely used dataset for MR and HD tasks, as it provides annotations for both tasks. This dataset marked the beginning of a trend where these two tasks are increasingly studied together. It includes more than 10,000 daily vlogs and news videos with text queries. Our main experiments were conducted on this dataset, and we provide comprehensive comparisons with other methods on it. Charades-STA~\cite{gao2017tall} and TACoS~\cite{regneri2013tacos} were used to evaluate the model's performance on MR, containing daily activities and cooking-related content, respectively. The other two datasets, TVSum~\cite{song2015tvsum} and YouTube-HL~\cite{sun2014ranking}, are sports-related and were used for HD evaluation.
\subsection{Evaluation Metrics}
We follow previous works~\cite{lei2021detecting, liu2024tuning, jiang2024llmepet} and adopt consistent evaluation metrics: R1@X, mAP, and mIoU for MR, and mAP and Hit@1 for HD. Specifically, R1@X stands for ``Recall 1 at X", which refers to selecting the predicted moment with the highest confidence score and checking whether its IoU with any ground truth moment exceeds the threshold X. If it does, the prediction is considered positive, and R@X is then calculated at thresholds $X \in \{0.3, 0.5, 0.7\}$. For MR, mAP serves as the primary metric, representing the mean of the average precision (AP) across all queries at thresholds [0.5:0.05:0.95]. We also use mean Intersection over Union (mIoU) to evaluate the average overlap between predicted moments and ground truth segments. For HD, mAP remains the key evaluation metric, with Hit@1 used to assess the hit ratio for the highest-scored clip.
\begin{table}
\small
\begin{tabular}{lcccc}
\toprule
\multirow{2}{*}{\textbf{Method}} & \multicolumn{2}{c}{test} & \multicolumn{2}{c}{val} \\ 
\cmidrule(lr{0.25em}){2-3} \cmidrule(lr{0.25em}){4-5} 
                                 & mAP         & HIT@1      & mAP        & HIT@1      \\ 
\hline
M-DETR~\cite{lei2021detecting}          & 35.69       & 55.60      & 35.65      & 55.55      \\
UMT~\cite{liu2022umt}                   & 38.18       & 59.99      & 39.85      & 64.19      \\
QD-DETR~\cite{moon2023qddetr}           & 38.94       & 62.40      & 39.13      & 63.03      \\
UniVTG~\cite{lin2023univtg}             & 38.20       & 60.96      & 38.83      & 61.81      \\
EaTR~\cite{jang2023eatr}                & -           & -          & 37.15      & 58.65      \\
CG-DETR~\cite{moon2023cgdetr}           & 40.33       & \underline{66.21}      & 40.79      & 66.71      \\
$R^2$-Tuning~\cite{liu2024tuning}       & 40.75       & 64.20      & 39.45      & 64.13      \\
LLMEPET~\cite{jiang2024llmepet}         & 40.33       & 65.69      & 40.52      & 65.03      \\ 
\cmidrule(lr{0.25em}){1-5}
\textbf{FlashVTG} \text{(Ours)}   & \underline{41.07}   & 66.15      & \underline{41.39}      & \underline{67.61}      \\
\textbf{FlashVTG}\textsuperscript{\dag} \text{(Ours)}   & \textbf{44.09}       & \textbf{71.01}      & \textbf{44.15}      & \textbf{72.90}      \\ 
\bottomrule
\end{tabular}%
\caption{Experimental results for Highlight detection on the QVHighlights~\cite{lei2021detecting}. All models used the same backbone, except for $R^2$-Tuning~\cite{liu2024tuning}, which used CLIP~\cite{radford2021clip} as the backbone, and FlashVTG marked with ``\dag", which used InternVideo2~\cite{wang2024internvideo2} as the backbone.}
\label{tab:qv-HD}
\vspace{-2em}
\end{table}
\begin{table*}[ht]
    \centering
    \begin{minipage}[t]{0.6\textwidth}
    \setlength{\tabcolsep}{0.5mm}{
    \centering
    \small
    \begin{tabular}{lccccccccccc}
    \toprule
    \textbf{Method} & \textbf{VT} & \textbf{VU} & \textbf{GA} & \textbf{MS} & \textbf{PK} & \textbf{PR} & \textbf{FM} & \textbf{BK} & \textbf{BT} & \textbf{DS} & \textbf{Avg.} \\ 
    \hline
    LIM-S \cite{xiong2019less} & 55.9 & 42.9 & 61.2 & 54.0 & 60.4 & 47.5 & 43.2 & 66.3 & 69.1 & 62.6 & 56.3 \\
    Trailer \cite{wang2020learning} & 61.3 & 54.6 & 65.7 & 60.8 & 59.1 & 70.1 & 58.2 & 64.7 & 65.6 & 68.1 & 62.8 \\
    SL-Module \cite{xu2021cross} & 86.5 & 68.7 & 74.9 & 86.2 & 79.0 & 63.2 & 58.9 & 72.6 & 78.9 & 64.0 & 73.3 \\
    PLD \cite{wei2022learning} & 84.5 & 80.9 & 70.3 & 72.5 & 76.4 & 87.2 & 71.9 & 74.0 & 74.4 & 79.1 & 77.1 \\
    UniVTG \cite{lin2023univtg} & 83.9 & 85.1 & 89.0 & 80.1 & 84.6 & 81.4 & 70.9 & \underline{91.7} & 73.5 & 69.3 & 81.0 \\ 
    $R^2$-tunning~\cite{liu2024tuning} & 85.0 & 85.9 & 91.0 & 81.7 & \textbf{88.8} & 87.4 & \underline{78.1} & 89.2 & \underline{90.3} & 74.7 & 85.2 \\
    LLMEPET~\cite{jiang2024llmepet} & \textbf{90.8} & \underline{91.9} & \textbf{94.2} & \textbf{88.7} & 85.8 & \underline{90.4} & \textbf{78.6} & \textbf{93.4} & 88.3 & 78.7 & \textbf{88.1} \\ 
    \cmidrule(r{0.25em}){1-12}
    MINI-Net\textsuperscript{\dag}~\cite{hong2020mini} & 80.6 & 68.3 & 78.2 & 81.8 & 78.1 & 65.8 & 57.8 & 75.0 & 80.2 & 65.5 & 73.2 \\
    TCG\textsuperscript{\dag}~\cite{ye2021temporal} & 85.0 & 71.4 & 81.9 & 78.6 & 80.2 & 75.5 & 71.6 & 77.3 & 78.6 & 68.1 & 76.8 \\
    Joint-VA\textsuperscript{\dag}~\cite{badamdorj2021joint} & 83.7 & 57.3 & 78.5 & 86.1 & 80.1 & 69.2 & 70.0 & 73.0 & \textbf{97.4} & 67.5 & 76.3 \\
    CO-AV\textsuperscript{\dag}~\cite{li2022probing} & \textbf{90.8} & 72.8 & 84.6 & 85.0 & 78.3 & 78.0 & 72.8 & 77.1 & 89.5 & 72.3 & 80.1 \\
    UMT\textsuperscript{\dag}~\cite{liu2022umt} & 87.5 & 81.5 & 88.2 & 78.8 & 81.4 & 87.0 & 76.0 & 86.9 & 84.4 & \underline{79.6} & 83.1 \\
    \cmidrule(r{0.25em}){1-12}
    \textbf{FlashVTG} \text{(Ours)} & \underline{88.32} & \textbf{94.33} & \underline{91.5} & \underline{87.7} & \underline{87.08} & \textbf{91.12} & 74.7 & \textbf{93.4} & \underline{90.3} & \textbf{81.7} & \underline{88.0} \\ 
    \bottomrule
    \end{tabular}%
    }
    \caption{Highlight detection results (Top-5 mAP) on TV-Sum~\cite{song2015tvsum} across different class. ``\dag" denotes the methods that utilize the audio modality.}
    \label{tab:TvSum}
\end{minipage}
\begin{minipage}[t]{0.38\textwidth}
    \setlength{\tabcolsep}{0.5mm}{
    \centering
    \small
    \begin{tabular}{lccccccc}
    \toprule
    \textbf{Method}  & \textbf{Dog}  & \textbf{Gym.} & \textbf{Par.} & \textbf{Ska.} & \textbf{Ski.} & \textbf{Sur.} & \textbf{Avg.} \\ \hline
    RRAE~\cite{yang2015unsupervised}            & 49.0 & 35.0 & 50.0 & 25.0 & 22.0 & 49.0 & 38.3 \\
    GIFs~\cite{gygli2016video2gif}              & 30.8 & 33.5 & 54.0 & 55.4 & 32.8 & 54.1 & 46.4 \\
    LSVM~\cite{sun2014ranking}                       & 60.0 & 41.0 & 61.0 & 62.0 & 36.0 & 61.0 & 53.6 \\
    LIM-S~\cite{xiong2019less}                  & 57.9 & 41.7 & 67.0 & 57.8 & 48.6 & 65.1 & 56.4 \\
    SL-Module~\cite{xu2021cross}                        & 70.8 & 53.2 & 77.2 & 72.5 & 66.1 & 76.2 & 69.3 \\
    QD-DETR~\cite{moon2023qddetr}                       & 72.2 & \textbf{77.4} & 71.0 & 72.7 & 72.8 & 80.6 & 74.4 \\
    LLMEPET~\cite{jiang2024llmepet}             & \underline{73.6} & 74.2 & 72.5 & \textbf{75.3} & \textbf{73.4} & 82.5 & \underline{75.3} \\ 
    \cmidrule(r{0.25em}){1-8}
    MINI-Net\textsuperscript{\dag}~\cite{hong2020mini} & 58.2 & 61.7 & 70.2 & 72.2 & 58.7 & 65.1 & 64.4 \\
    TCG\textsuperscript{\dag}~\cite{ye2021temporal}  & 55.4 & 62.7 & 70.9 & 69.1 & 60.1 & 59.8 & 63.0 \\
    Joint-VA\textsuperscript{\dag}~\cite{badamdorj2021joint} & 64.5 & 71.9 & \underline{80.8} & 62.0 & \underline{73.2} & 78.3 & 71.8 \\
    UMT\textsuperscript{\dag}~\cite{liu2022umt}        & 65.9 & 75.2 & \textbf{81.6} & 71.8 & 72.3 & \underline{82.7} & 74.9 \\
    UniVTG\textsuperscript{\dag}~\cite{lin2023univtg}   & 71.8 & \underline{76.5} & 73.9 & 73.3 & \underline{73.2} & 82.2 & 75.2 \\ 
    \cmidrule(r{0.25em}){1-8}
    \textbf{FlashVTG} \text{(Ours)}            & \textbf{76.5} & 76.1 & 69.4 & \underline{74.1} & 73.1 & \textbf{83.0} & \textbf{75.4} \\ 
    \bottomrule
    \end{tabular}%
    }
    \caption{Highlight detection performances (mAP) on YouTube-HL~\cite{sun2014ranking} across different class. ``\dag'' indicates the usage of audio modality.}
    \label{tab:youtube-hl}
\end{minipage}
\vspace{-1em}
\end{table*}

\subsection{Implementation Details}
As in previous methods~\cite{moon2023qddetr, liu2022umt, lei2021detecting}, we primarily used video and text features extracted by CLIP~\cite{radford2021clip} and SlowFast~\cite{feichtenhofer2019slowfast} across all five datasets for a fair comparison. To further verify the generalizability of our model, we incorporated video and text features extracted by InternVideo2~\cite{wang2024internvideo2} and LLaMA~\cite{touvron2023llama} for QVHighlights and Charades-STA, as well as features extracted by VGG~\cite{simonyan2014vgg} and GloVe~\cite{pennington2014glove} for Charades-STA. All feature dimensions were set to 256. The number of attention heads in the Feature Fusion Module was set to 8, with $K=4, 5$ layers in temporal feature layering, and the number of 2D convolutional layers in the score head was set to 2. AdamW was used as the optimizer, and the NMS threshold during inference was set to 0.7. Training was conducted on a single Nvidia RTX 4090 GPU, taking approximately one and a half hours for 150 epochs on QVHighlights.

\subsection{Comparison Results}
\label{subsec:comparision results}
The comparison of MR experimental results is shown in Tables~\ref{tab:qv_mr}, \ref{tab:tacos}, \ref{tab:Charades-STA}, \ref{tab:short-mAP}, where FlashVTG achieved state-of-the-art (SOTA) performance on nearly all metrics. Table~\ref{tab:qv_mr}, \ref{tab:short-mAP} presents the MR results on the QVHighlights~\cite{lei2021detecting}. Table~\ref{tab:qv_mr} demonstrates that combining FlashVTG with a newly selected backbone~\cite{wang2024internvideo2} resulted in a significant performance improvement, evidenced by a 5.8\% increase in mAP on the test set. Even when using the same backbone, mAP improved by 1.4\%. Table~\ref{tab:short-mAP} shows that FlashVTG improved the mAP of short moment retrieval to 125\% of the previous SOTA method.These performance gains on one of the most widely used datasets demonstrate the effectiveness of FlashVTG compared to contemporaneous approaches.

Tables~\ref{tab:tacos} and \ref{tab:Charades-STA} present experimental results on two other MR datasets, TaCos and Charades-STA. Similarly, FlashVTG significantly outperformed previous methods on almost all metrics, achieving either SOTA or the second-highest performance. On Charades-STA, performance improvements were observed across all three different backbones, further validating the robustness of FlashVTG.

The comparison of HD experimental results is shown in Tables~\ref{tab:qv-HD}, \ref{tab:TvSum}, \ref{tab:youtube-hl}. FlashVTG achieved SOTA performance on both QVHighlights and YouTube-HL, and outperformed all methods that used additional audio features on TVSum, reaching performance comparable to the SOTA. This indicates that FlashVTG can deliver excellent performance even with relatively smaller datasets.

\begin{table}
    \centering
    \small
    \begin{tabular}{@{}lcc@{}}
    \toprule
    \textbf{Method}   & MR-short-mAP & mAP   \\ \midrule
    UMT~\cite{liu2022umt}      & 5.02         & 38.59 \\
    UniVTG~\cite{lin2023univtg}   & 8.64         & 36.13 \\
    CG-DETR~\cite{moon2023cgdetr}  & 10.58        & 44.93 \\
    LLMEPET~\cite{jiang2024llmepet}  & 10.96        & 46.24 \\
    $R^2$-Tuning~\cite{liu2024tuning}       & \underline{12.62}        & \underline{47.86} \\ \midrule
    \textbf{FlashVTG} (Ours) & \textbf{15.73}        & \textbf{49.85} \\
    \bottomrule
    \end{tabular}%
    \caption{Performance comparison for short moment (\textless 10s) retrieval on QVHighlights~\cite{lei2021detecting}. All methods, except $R^2$-tunning~\cite{liu2024tuning}, use SlowFast~\cite{feichtenhofer2019slowfast} + Clip~\cite{radford2021clip} as the backbone.}
    \label{tab:short-mAP}
\vspace{-1.5em}
\end{table}

\subsection{Ablation Study}
\label{sunsec: ablation study}
We conducted an ablation study on the QVHighlights validation set to verify the effectiveness of FlashVTG. This dataset is currently the most widely used VTG benchmark and supports both MR and HD tasks, making it the most suitable for our research. We used the original single-layer feature and a method that generates confidence scores based solely on single predicted moments as our baseline.

\textbf{Effect of different components.} As shown in Table~\ref{tab:component ablation}, we studied the effects of the Temporal Feature Layering and Adaptive Score Refinement modules. It can be observed that adding the TFL module to the baseline improved performance in both the MR and HD tasks, with the mAP for MR increasing by nearly 6\%. Building on this, we further incorporated the ASR module to refine the confidence scores of the predicted moments. The improvements in MR-mAP, R@0.5, and R@0.7 indicate that the overall precision was maintained while enhancing Recall at 1, meaning the recall for the first predicted moment improved. Additionally, using the saliency score label from the HD task as a supervision signal indirectly boosted HD performance, with mAP increasing by 0.5\% and Hit@1 improving by 1\%.
\begin{table}
\centering
\small
\begin{tabular}{@{}c|l|ccc@{}}
\toprule
\multirow{2}{*}{\textbf{Component}} & \multicolumn{1}{l|}{\textbf{+ TFL Module}} &  & \textbf{\checkmark} & \textbf{\checkmark} \\
                                  & \multicolumn{1}{l|}{\textbf{+ ASR Module}} &  & & \textbf{\checkmark} \\ \midrule
\multirow{5}{*}{\textbf{Metrics}} & MR-R1@0.5    & 68.26 & {\ul 72.39} & \textbf{73.10}       \\
                                  & MR-R1@0.7    & 51.35 & {\ul 56.19} & \textbf{57.29}      \\
                                  & MR-mAP       & 46.84 & {\ul 52.47} & \textbf{52.84}      \\
                                  & HL-mAP       & 42.20 & {\ul 43.63} & \textbf{44.15}      \\
                                  & HL-Hit1      & 69.23 & {\ul 71.81} & \textbf{72.90}       \\ \bottomrule
\end{tabular}%
\caption{Ablation study on different components. TFL stands for Temporal Feature Layering, ASR stands for Adaptive Score Refinement.}
\label{tab:component ablation}
\vspace{-1.5em}
\end{table}

\vspace{-2em}
\section{Conclusion}
\label{sec:conclusion}
This paper introduced FlashVTG, a novel architecture for Video Temporal Grounding tasks. The proposed Temporal Feature Layering and Adaptive Score Refinement modules improve the retrieval and selection of more accurate moment predictions across varying lengths. Extensive experiments on five VTG benchmarks demonstrate that FlashVTG consistently outperforms state-of-the-art methods in both Moment Retrieval and Highlight Detection, achieving substantial improvements. These results validate the robustness and effectiveness of our approach, establishing FlashVTG as a leading solution for VTG tasks.
\vspace{-1em}
\section*{Acknowledgment}
This work is supported by Australian Research Council (ARC) Discovery Project DP230101753.

{\small
\bibliographystyle{ieee_fullname}
\bibliography{egbib}
}

\end{document}